\documentclass[journal]{IEEEtran}
\usepackage{graphicx}
\usepackage{amsmath,amssymb} 
\usepackage{color}
\usepackage{epstopdf}
\usepackage{algorithm}
\usepackage{algpseudocode}
\usepackage{epsfig}
\usepackage{url}
\usepackage{caption}
\usepackage{multirow,array}
\usepackage{stackengine}
\usepackage{float}

\begin{document}
\title{Flow-free Video Object Segmentation}
\author{\IEEEauthorblockN{Aditya Vora\IEEEauthorrefmark{1} and
Shanmuganathan Raman\IEEEauthorrefmark{2}\\}
\IEEEauthorblockA{Electrical Engineering,
Indian Institute of Technology Gandhinagar, Gujarat, India, 382355\\
Email: \IEEEauthorrefmark{1}aditya.vora@iitgn.ac.in,
\IEEEauthorrefmark{2}shanmuga@iitgn.ac.in}}

\maketitle

\begin{abstract}
Segmenting foreground object from a video is a challenging task because of the large deformations of the objects, occlusions, and background clutter. In this paper, we propose a frame-by-frame but computationally efficient approach for video object segmentation by clustering visually similar generic object segments throughout the video. Our algorithm segments various object instances appearing in the video and then perform clustering in order to group visually similar segments into one cluster. Since the object that needs to be segmented appears in most part of the video, we can retrieve the foreground segments from the cluster having maximum number of segments, thus filtering out noisy segments that do not represent any object. We then apply a track and fill approach in order to localize the objects in the frames where the object segmentation framework fails to segment any object. Our algorithm performs comparably to the recent automatic methods for video object segmentation when benchmarked on DAVIS dataset while being computationally much faster. 
\end{abstract}


\IEEEpeerreviewmaketitle

\section{Introduction}

\label{sec:introduction}

Video object segmentation is the task of separating a foreground object from the background in a video. Previous algorithms for video object segmentation make use of optical flow in order to localize foreground object in each frame of the video (\cite{ochs2011object,brox2010object,papazoglou2013fast,
zhang2013video,lee2011key}). However, computing optical flow in each frame is a computationally expensive task especially when the spatial resolution of the video frame is large. Also, because of the inaccuracies of the optical flow estimation at the boundaries of the object, it becomes a challenging task to model the temporal consistency of the object segmentation and this leads to error propagation over time. As a result, it becomes practically infeasible to scale these algorithms to longer video sequences with a larger spatial resolution. Thus, instead of computing dynamic cues of object proposals from the optical flow, we perform a non-parametric clustering of these object proposals which will enable us to cluster the foreground object proposals. Because of the fact that the object that needs to be segmented appears throughout the frames consistently irrespective of deformations and occlusions, we can obtain the foreground segments by selecting the cluster that has the most number of object proposals. Moreover, when we consider the clustering process in our algorithm, the features are computed using a convolutional neural network (CNN) \cite{simonyan2014very} which helps us to restrict the spatial resolution of the segment to the network architecture (i.e., 224$\times$224). Because of this, the clustering process takes almost constant time irrespective of the spatial resolution of the original video. Due to this advantage, our algorithm can scale up to longer video sequences with larger spatial resolutions. In order to localize the object properly in the frames, where either the detection framework might have failed to detect the segment or the presence of noise in the clustering process, we employ a track and fill approach in order to estimate the segments in those frames. First, we track the object in the frames where the segmentation is not available in order to get a loosely bounded window around the object. Then, we compute the average mask of the segments that are already detected in the nearest set of frames. We utilize this average mask to initialize the GrabCut algorithm \cite{rother2004grabcut}, using an approach similar to that described by Kuettel \emph{et al.} \cite{kuettel2012figure}.
\vspace{3mm} \\
In this work, the following major contributions are made: 
\begin{itemize}
\item Given a video sequence, our algorithm discovers, segments as well as clusters generic objects, where the category of the objects appearing in the video is assumed to be unknown. 
\item We propose a flow-free algorithm for video object segmentation that instead makes use of the cluster information in order to obtain localized segments throughout the frames of the video.
\item  We extend the segmentation transfer on images \cite{kuettel2012figure} to videos in order to obtain temporally consistent segments.
\end{itemize}

\section{Related Work}
\label{sec:related work}
\textbf{Fully automatic approaches:} Fully automatic methods for video object segmentation assume no user input during runtime. They are grouped into several categories. First, we have the algorithms which oversegment the frames of the video into space-time superpixels with similar appearance and motion \cite{grundmann2010efficient,xu2012evaluation}. However, the goal of our work is different from these approaches. We focus on generating an accurate spatio-temporal tube of binary masks which are well aligned around the boundaries of the foreground object appearing in the video. Other category of algorithms exploits long term motion cues in the form of dense trajectories which are derived using dense optical flow in order to get the final object level segmentation \cite{brox2010object,ochs2011object}. However, because of the assumption that the object moves with only a single type of translation, these technique sometimes fail to give proper results with non-rigid objects. Only recently, methods based on finding region proposals were proposed \cite{lee2011key,zhang2013video}. All these methods generate thousands of object-like space-time segments and try to rank them based on some static or dynamic cues and after selecting some best ranked segments they try to associate these segments across the space and time of the video. In \cite{faktor2014video}, the authors tries to expand the concept of saliency object detection to videos. Although it is a unsupervised approach it is based on the assumption that the object motion is dissimilar from the surroundings. In \cite{papazoglou2013fast}, a fast algorithm for video object segmentation was proposed which tries to localize the dynamic object in the video based on the motion boundaries computed from the optical flow.
\\
\textbf{Supervised approaches:} Supervised approaches for video object segmentation requires the user to manually annotate some of the frames in the video and the algorithms then propagates the information to the rest of the frames \cite{bai2009video,wang2012probabilistic}. Other than this, some of the other algorithms require user to annotate the first frame of the video with the corresponding segmentation and then track the object in the rest of the video \cite{chockalingam2009adaptive,TsaiBMVC10}. Nicolas \emph{et al.} recently proposed a semi-automatic algorithm to video object segmentation that operates in the bilateral space achieving good accuracy \cite{Maerki_CVPR_2016}. These algorithms use motion in order to primarily propagate information to the other frames of the video. Caelles \emph{et al.} proposed an approach that uses annotation of the first frame and then using the generic semantic information of the fully convolutional neural network transfers the foreground segment information to the rest of the frames \cite{Cae+17}.

\section{Proposed Approach}
\begin{figure}[h!]
\begin{center}
   \includegraphics[width=1.0\linewidth]{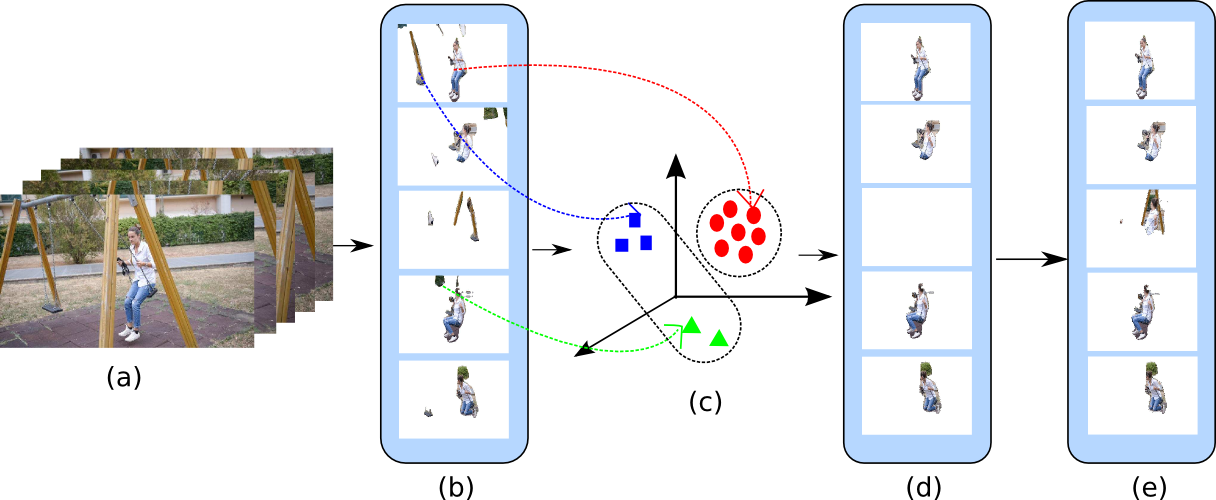}
\end{center}
   \caption{\textbf{Strategy for efficient video object segmentation.} (a) Input video. (b) Top-5 region proposals of each frame are unioned. (c) Mean Shift Clustering on the features extracted from the FC6 layer of VGG-16 CNN model. Cluster with most segments is shown in red which represents the foreground segment, while non-object segments are shown in blue and green. (d) Result after clustering. (e) Final segmentation after track and fill.}
\label{fig:pipeline}
\end{figure}

Given a video, we need to discover and segment the foreground object appearing in the video. The overview of our approach is shown in Fig. \ref{fig:pipeline}. 

\subsection{Extracting object like candidates}

\begin{figure}[h!]
\begin{center}
   \stackunder{\includegraphics[width=0.24\linewidth]{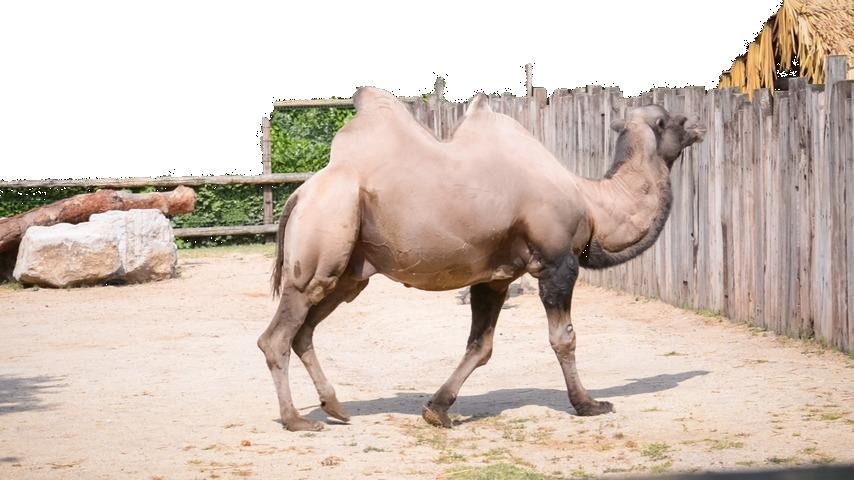}}{$(a)$}
   \stackunder{\includegraphics[width=0.24\linewidth]{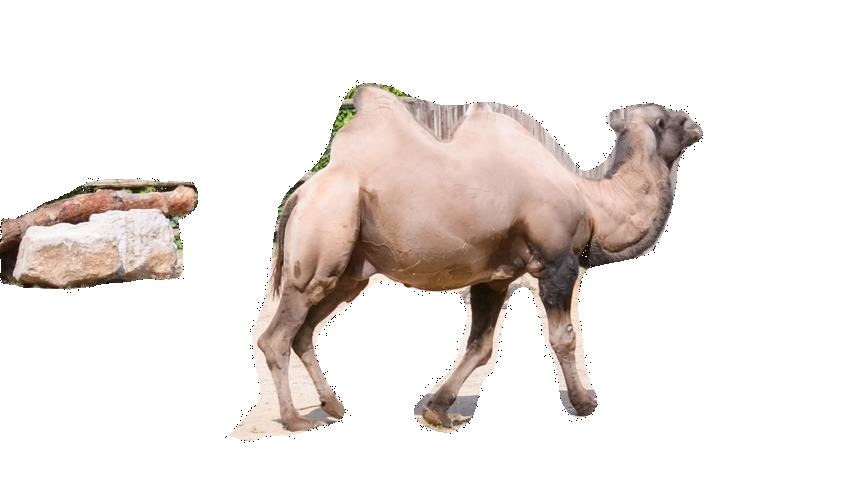}}{$(b)$}
   \stackunder{\includegraphics[width=0.24\linewidth]{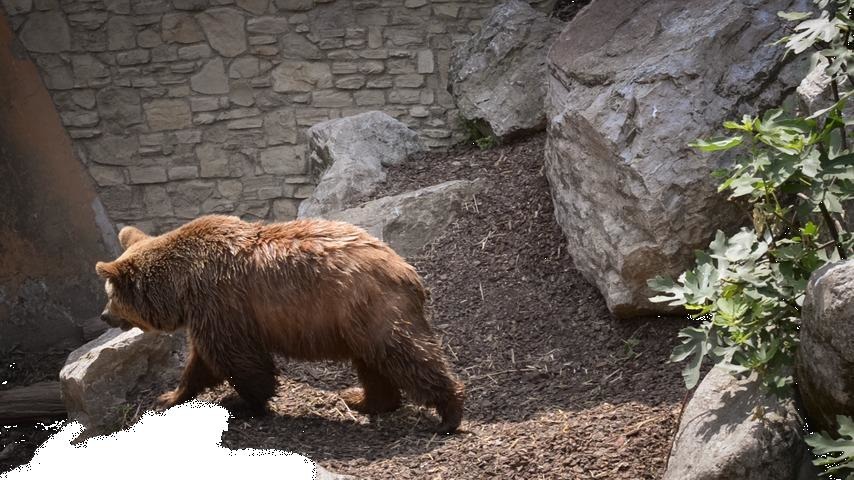}}{$(c)$}
   \stackunder{\includegraphics[width=0.24\linewidth]{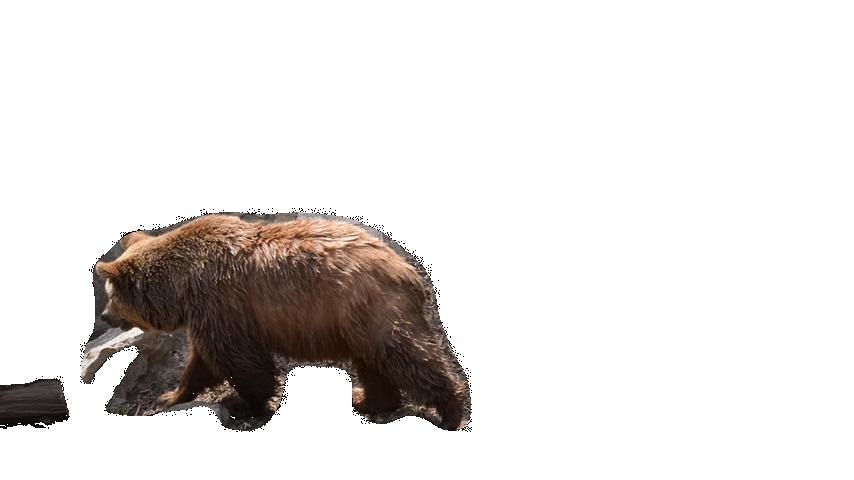}}{$(d)$}
\end{center}
   \caption{$(a),(c)$: Union of top-5 proposals generated from \cite{endres2010category}. $(b),(d)$: Union of top-5 proposals generated from \cite{pinheiro2016learning}.}
\label{fig:proposals}
\end{figure}

Several previous video segmentation algorithms \cite{lee2011key,
zhang2013video} compute category independent bag of regions in each frame and rank them such that the top ranked region contains the segments of different objects. The most popular technique for generating these regions was proposed by Endres \emph{et al.} \cite{endres2010category}. This technique however has the following drawbacks. (i) It generates thousands of regions for each frame of the video. As a result, it becomes extremely challenging to associate thousands of these regions from different objects while maintaining temporal connections for each of them across all the sequences. (ii) It is computationally too expensive. It takes about few minutes/frame in order to generate candidate regions in each frame. (iii) Top most proposals are always not aligned around the object. As shown in the Fig. \ref{fig:proposals} (a),(c) which the is union of top-5 regions generated using the technique in \cite{endres2010category}, it almost covers the entire image with its top most regions which requires further processing in order to improve the spatial accuracy of the segment. 

Deepmask is a discriminative CNN which is trained to perform object instance segmentation over the image given as input \cite{pinheiro2015learning}. Sharpmask on the other end is another object instance segmentation model which is built over deepmask to generate refined object segments \cite{pinheiro2016learning}. It first generates a coarse mask with a bottom-up approach using deepmask and then fuses appearance information from the fine layers of the network using a refinement module in order to obtain spatially accurate segments of the object instances in each test image. This bottom-up/top-down architecture for segmentation has high recall capability using fewer masks which can be seen in Fig. \ref{fig:proposals} (b),(d). It gives reasonably good localization of the object by selecting very few region proposals. We use the publicly available pre-trained model of sharpmask (\cite{pinheiro2016learning,pinheiro2015learning}). As compared to the technique in \cite{endres2010category}, the segmentation mask that is generated is properly aligned to the object. We select top-$k$ region proposals from this model so that we can obtain the preliminary mask of that particular frame. Selecting appropriate value of $k$ is necessary as it will affect the results in the following ways: (1) Selecting lower value of $k$ will not be able to localize the object properly and thus will not lead to spatially accurate segments. (2) Selecting higher value of $k$ will lead to more non-object like proposals in each frame and thus will affect the clustering accuracy. As a result, we select $k=5$, which we keep unchanged throughout our experiments. Thus taking the union of top-$5$ region proposals from this model could easily result into a preliminary mask for that particular frame. 

Given a video $\mathcal{V}$ with a set of $N$ frames, for each frame $\mathcal{V}_{i} \in \mathcal{V}$, we process them using the pre-trained model of sharpmask \cite{pinheiro2016learning}. This model is applied to the entire set of frames efficiently and is able to generate a set of region proposals $\mathcal{B}_{i} = \{b_{i,1}, b_{i,2},..., b_{i,k}\}$ for each frame. Here we had selected $k=5$ and the masks are binarized with a threshold of $0.2$. Then in order to obtain a preliminary mask $\mathcal{S}_{i}$ for each frame $i$, we take the union of all the $k$ region proposals $S_{i} = b_{i,1} \cup b_{i,2}...\cup b_{i,k}$ corresponding to that frame which are generated by the segmentation framework. As a result, we now obtain $N$ preliminary masks $S = \{S_{1},S_{2},...,S_{N}\}$ corresponding to each frame of the video.

\subsection{Clustering of visually similar generic objects}

Preliminary segmentation mask obtained by taking the union of top-5 region proposals generated from the sharpmask consists of region proposals covering the foreground object as well as some noisy region proposals that do not represent any object. This can be seen in Fig. \ref{fig:pipeline}(b). In order to filter out these non-object like segments and to localize the foreground segment well throughout all frames, we extract a $4096$ dimensional descriptor from the FC6 layer of VGG-Net \cite{simonyan2014very} for each region proposal in the frame. This gives us a general feature level representation of each proposal appearing in each frame of the video. We then normalize each feature vector using the $L_{2}$ norm. After computing these features for each region proposal, we do a mean shift clustering on the computed features. The main reason for using mean shift clustering is the fact that it does not make any assumption about the predefined shape of the data clusters and it can automatically decide over the optimal number of clusters based on the window size. After clustering, since the similarity between the foreground segments is high throughout all the frames, all the foreground segments will be accommodated in one cluster and all other noisy proposals will be distributed in different clusters. In order to select the cluster having foreground segments automatically we put a threshold of $0.6$ on the minimum size of the cluster that the mean shift algorithm would select. As a result of this, the object that needs to be segmented will automatically get selected and other noisy components will get filtered. 

\subsection{Tracking and Filling}
Sometimes there is a chance that the object that needs to be segmented remains undetected in few frames of the video as it gets missed by the segmentation detection framework. It might also be possible that some object segments of few frames are lost due to the inefficiency of the clustering process. Some techniques use segmentation tracking strategies in order to fill out these regions. We propose a technique in order to properly localize objects in undetected frames which is based on segmentation transfer approach. Segmentation transfer techniques are used to estimate the segmentation on a new test image from the set of images for which the segmentation masks are already known. Our idea of track and fill is based on the segmentation transfer technique proposed by \cite{kuettel2012figure}. 

\begin{figure}[h!]
\begin{center}
   \includegraphics[width=1.0\linewidth]{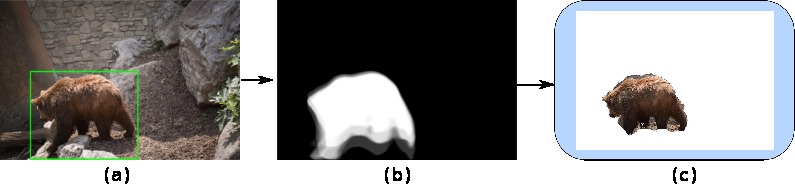}
\end{center}
   \caption{\textbf{Approach to Track and Fill.} (a) Window for segmentation transfer obtained from tracking on undetected frame $X$. (b) Soft segmentation mask $\mathcal{M}$ of $p$ nearest frames. Here $p=10$. (c) Final mask obtained from GrabCut.} 
\label{fig:segment transfer}
\end{figure}

We first discover the frame where the segment is undetected. Let that frame be $X$. We then find the nearest set of $p$ frames $N_{p}$ from the generated object cluster. These selected set of frames are then used to compute the soft segmentation mask $\mathcal{M}$. In order to do so, we first localize the region where we need to transfer the average segmentation mask. For this, we track the region from the nearest frame where the segment is present to the frame $X$. We use compressive tracking for this purpose \cite{zhang2012real}. After getting this bounding box window on the test image $X$, we resize all the $p$ windows of $N_{p}$ containing their respective segments to the size of the bounding box region on the test image. These resized windows are denoted as $W_{p}$. Then these windows are pixel-wise averaged in order to obtain the soft segmentation mask $\mathcal{M}$. We use this soft-segmentation mask $\mathcal{M}$ in order to initialize the GrabCut algorithm. 

We formulate the problem as an energy minimization problem which can be solved using graph-cut. Given a test image $X$, neighbouring windows $W_{p}$ of all neighbouring frames $N_{p}$ and soft segmentation mask $\mathcal{M}$, we would like to label each pixel $x_{i} \in X$ as foreground $(c_{i}=1)$ or background $(c_{i}=0)$ which will result into labelling $C$ of the entire frame $X$. The energy function that needs to be minimized is given by $E(C) = \sum_{i} \phi_{i}(r_{i}) + \sum_{ij\in \epsilon} \psi_{ij}(c_{i},c_{j})$. It has two terms, first is the unary potential $\phi_{i}$ and second is the pairwise potential $\psi_{ij}$. Unary potential $\phi_{i}$ is the data term which is obtained from the soft segmentation mask $\mathcal{M}$ and is given as $\phi_{i}(c_{i}) = -\log P(x_{i}|c_{i})-\log Q_{i}(c_{i})$, where $i$ is the index of the pixel in the test image $X$. The two terms in the unary potential are $P(x_{i}|c_{i})$ which is the appearance based model and $Q_{i}(c_{i})$ is the location based model. Appearance based model uses Gaussian mixture model (GMM) on the RGB color space to model the foreground and the background. We fit two GMMs, $P_{1}$ for foreground and $P_{0}$ for background. Location based model $Q$ defines in what location of the image is likely to contain foreground/background. For foreground, it is given as $Q_{i}(c_{i}=1) = M(i)$ and for background, it is given as $Q_{i}(c_{i}=0) = 1 - M(i)$. The pairwise potential $\psi_{ij}$ is the standard penalty term used in many standard graph cut algorithms which is given by $\psi_{ij}(c_{i},c_{j}) = \gamma d^{-1}(i,j)[c_{i} \neq c_{j}]e^{-\beta |x_{i}-x_{j}|^2}$. Here the neighbourhood is selected using $\epsilon$. Generally 8-neighbourhood is considered. It is designed to penalize neighbouring pixels to take different label values. Optimum labelling $C^{*}$ is obtained by minimizing the energy function $E(C)$ which can be obtained by max-flow technique described in \cite{boykov2004experimental}. Solving this graph-cut problem shall result into segmentation in the test image $X$. 

\section{Evaluation}

\textbf{Dataset:} We evaluate our algorithm on the recently released DAVIS Dataset which is the most challenging dataset for the benchmarking of video object segmentation algorithms \cite{perazzi2016benchmark}. It consists of 50 high quality video sequences with resolution of 480p and 1080p. We evaluate our algorithm on 480p video sequences as all other baseline algorithms which are compared with are evaluated on the same set of videos. 
\\
\textbf{Metric:} In order to measure the region similarity between the estimated segmentation masks and the ground truth, we make use of Jaccard index $\mathcal{J}$ which is measured as intersection-over-union between the estimated mask and the ground truth. Given the estimated mask $M$ and corresponding ground truth mask $G$, the Jaccard index $\mathcal{J}$ is computed as $\mathcal{J}=\frac{|M \cap G|}{|M \cup G|}$. We average out the Jaccard similarity score across all the frames of videos in the dataset in order to obtain the average intersection over union score which is represented as $\mathcal{J}$ Mean $(\mathcal{M})$ in Table \ref{table:evaluation}. The higher the value of $\mathcal{M}$ the better the algorithm is. We also compute the $\mathcal{J}$ Decay $(\mathcal{D})$ which quantifies the error propagation over time because of the inaccuracies in segmentation tracking and $\mathcal{J}$ Recall $(\mathcal{O})$ which measures the fraction of sequences scoring higher than a threshold. Recall should be higher and decay should be lower for a better performing algorithm.
\\
\textbf{Baselines:} We compare our technique with several other fully automatic state-of-the-art techniques for video object segmentation which are FST \cite{papazoglou2013fast}, SAL \cite{wang2015saliency}, KEY \cite{lee2011key}, MSG \cite{ochs2011object}, TRC \cite{fragkiadaki2012video}, CVOS \cite{taylor2015causal} and NLC \cite{faktor2014video}. Our algorithm has $\mathcal{J}$ Mean ($\mathcal{M}$) of \textbf{0.598}, Recall ($\mathcal{O}$) of \textbf{0.731}, and Decay ($\mathcal{D}$) of \textbf{0.018}. As it can be seen from Table \ref{table:evaluation}, our algorithm has comparable accuracy to NLC \cite{faktor2014video}, whereas it performs better than all remaining automatic algorithms. This is because of the following reasons: (1) Foreground estimation in our algorithm is done using sharpmask \cite{pinheiro2016learning} which uses structural information from various layers using a refinement module in order to estimate spatially accurate foreground segments. (2) Clustering of segments is done with VGG features \cite{simonyan2014very} which are discriminative in nature and thus able to filter out the noisy segments from the frames in an efficient manner. (3) Track and fill takes care of the temporal consistency of the segments throughout the frames of the video. Because of these reasons, our algorithm is able to perform well in terms of $\mathcal{J}$ Mean $(\mathcal{M})$ than other algorithms. Moreover, our algorithm has a lower $\mathcal{J}$ Decay $(\mathcal{D})$ than all other algorithms which shows that our method has least error propagation over time thus giving consistent segmentation results. This is because our algorithm computes segments frame-by-frame through the segmentation model because of which each frame is treated independently and thus leading to less error propagation through time. This is not the case with algorithms that make use of optical flow because of error in estimation of optical flow at the boundaries of the object which gets accumulated over time, resulting into a much larger decay in segmentation quality over time. 

\begin{table}[h]
    \centering
        \begin{tabular}{	|c|cc|cc|cc|}
		\hline
		\multicolumn{7}{|c|}{\textbf{Evaluation on DAVIS dataset}}\\
		\hline
		\textbf{Method} &      & \textbf{$\mathcal{J}$ Mean $\mathcal{M}$} &      & \textbf{$\mathcal{J}$ Recall $\mathcal{O}$} &      & \textbf{$\mathcal{J}$ Decay $\mathcal{D}$} \\
		\hline
		FST \cite{papazoglou2013fast} &      &  0.575 &      & 0.652 &      & 0.044 \\
		SAL \cite{wang2015saliency} &      & 0.426 &      & 0.386 &      & 0.084 \\
		KEY \cite{lee2011key} &      & 0.569 &      & 0.671 &      & 0.075 \\
		MSG \cite{ochs2011object} &      & 0.543 &      & 0.636 &      & 0.028 \\
		TRC \cite{fragkiadaki2012video} &      & 0.501 &      & 0.560 &      & 0.050 \\
		CVOS \cite{taylor2015causal} &      & 0.514 &      & 0.581 &      & 0.127 \\
		NLC \cite{faktor2014video} &      & {\textbf{0.641}} &      & 0.731 &      & 0.086 \\
		\hline
		Ours &      &  0.598 &      & {\textbf{0.731}} &      & {\textbf{0.018}} \\
		\hline
		\end{tabular}
		\caption{Evaluation of our algorithm on DAVIS Dataset \cite{perazzi2016benchmark}.} 	
    \label{table:evaluation}
\end{table}

Despite comparable accuracy, our algorithm is much more computationally efficient compared to the other fully automatic techniques. We perform the experiments in MATLAB environment on PC with Intel core i7 processor with NVIDIA Quadro K2200 GPU. Our algorithm takes $2$ sec/frame for object instance segmentation, $0.1$ sec/frame for feature extraction of all proposals in each frame and $3$ sec/frame for track and fill stage. The segmentation and feature extraction stage are fixed timing of our algorithm that are required for all the frames of the video. However, track and fill stage is the variable timing of our algorithm as actual computation time required by this stage depends on the number of frames in which the segment is undetected. We shall do a worst case analysis of our algorithm. If we assume that the track and fill stage is executed for all the frames of the video (i.e. segmentation framework fails to detect object in all the frames) then the average time required by our algorithm would be $5.1$ sec/frame. Compared to NLC technique which takes $12$ sec/frame as reported in the paper \cite{faktor2014video}, our technique is around $2.5 \times$ faster in the worst case. Moreover, they have reported the timings on the SegTrack dataset which have lower resolution videos compared to DAVIS dataset. FST technique, however computes segmentation within $0.5$ sec/frame \cite{papazoglou2013fast}. However, they assume that the optical flow and the superpixels are already available and thus are not included in the computation time calculation. As a result, if we add all these computational overheads, the overall computation time surely will increase significantly as computing optical flow for a video with 480p resolution is a computationally expensive task.

\begin{figure}[h]
\begin{center}
   \includegraphics[width=0.95\linewidth]{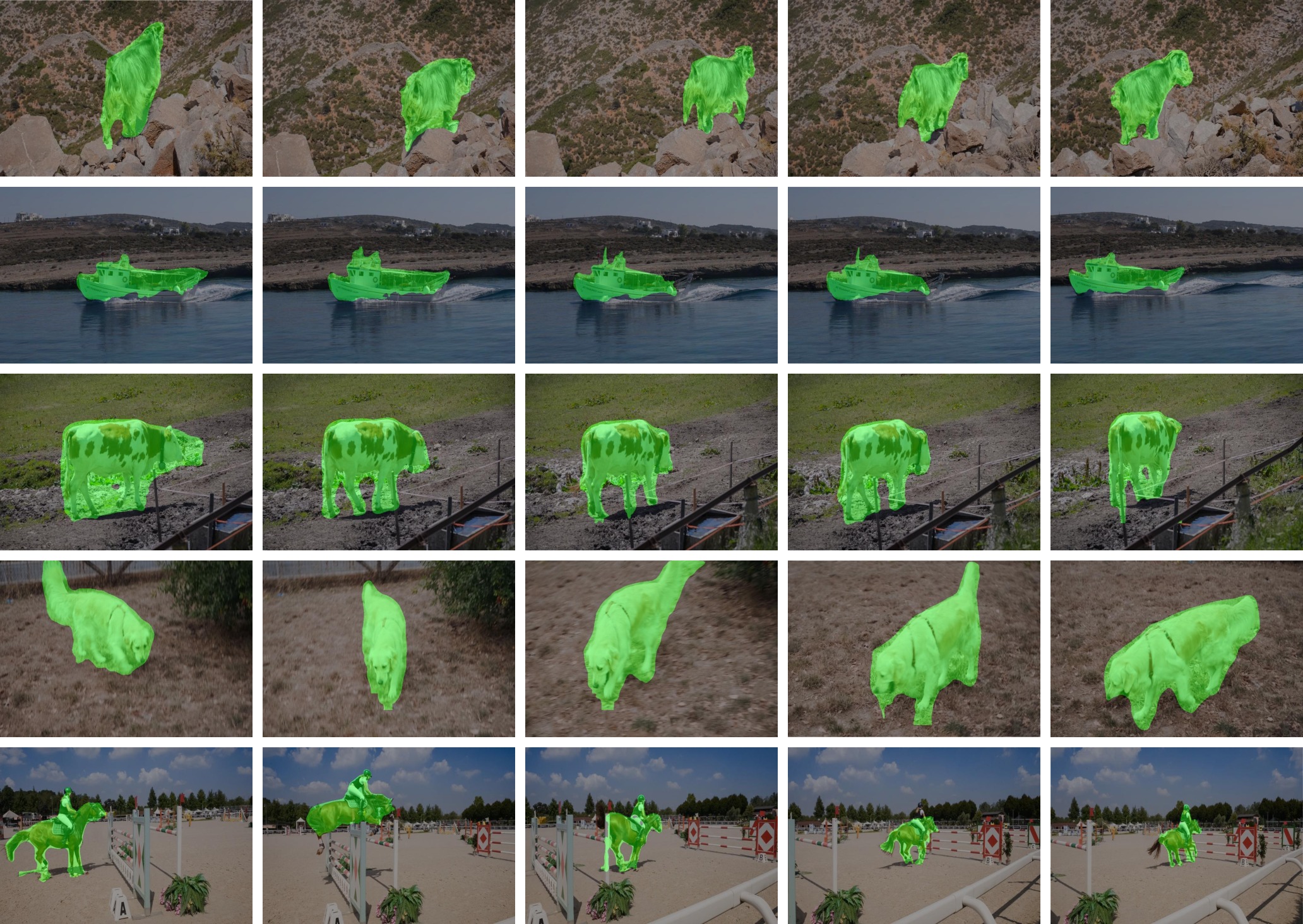}
\end{center}

   \caption{Example results on DAVIS dataset \cite{perazzi2016benchmark}.} 
\label{fig:final results}
\end{figure}

\section{Conclusion}
We have presented a new approach for segmenting generic objects in the video, that instead of using optical flow to localize the object instances in the video, makes use of the cluster information to do so. Results show comparable accuracy with respect to the state-of-the-art techniques. Our method performs better than most of the fully automatic techniques while being much faster than these algorithms. As our algorithm treats each frame of the video independently, it gives rise to spatially accurate and temporally consistent segments unlike algorithms which make use of optical flow. In future, we plan to explore extensions of this algorithm by intelligently including optical flow for localizing the foreground object in the initial frames of the video and then search for a similar object instance in the remaining frames of the video.  

\bibliographystyle{IEEEtran}
\bibliography{IEEEabrv,spl_submission}
\end{document}